\documentclass[accepted]{car2025} 
 
\usepackage[american]{babel}
\usepackage{natbib}
\usepackage{mathtools}   
\usepackage{booktabs} 
\usepackage{tikz} 
\usepackage{algorithm2e}
\usepackage{amsthm,amssymb} 
\usepackage{bbm} 

\newtheorem{theorem}{Theorem}
\newtheorem{lemma}{Lemma}
\newtheorem{corollary}{Corollary}
\theoremstyle{definition}

\newtheorem{example}{Example}
\theoremstyle{remark}

\makeatletter
\newcommand*{\centernot}{%
  \mathpalette\@centernot
}
\def\@centernot#1#2{%
  \mathrel{%
    \rlap{%
      \settowidth\dimen@{$\m@th#1{#2}$}%
      \kern.5\dimen@
      \settowidth\dimen@{$\m@th#1=$}%
      \kern-.5\dimen@
      $\m@th#1\not$%
    }%
    {#2}%
  }%
}
\makeatother

\newcommand{\val}{\mathrm{val}}

\newcommand{\graph}[1]{\mathcal{#1}}
\newcommand{\pa}[1]{\mathrm{Pa}(#1)}
\newcommand{\pag}[1]{\mathrm{Pa}_{\graph{G}}(#1)}
\newcommand{\Do}[1]{\mathrm{do}(#1)}
\newcommand{\dox}{\Do{\mathbf{x}}}
\newcommand{\doX}{\Do{\mathbf{X}=\mathbf{x}}}

\newcommand{\bracket}[1]{[\![#1]\!]}

\makeatletter
\def\munderbar#1{\underline{\sbox\tw@{$#1$}\dp\tw@\z@\box\tw@}}
\makeatother
\RestyleAlgo{ruled}

\title{Multilinear and Linear Programs for Partially Identifiable 
Queries in Quasi-Markovian Structural Causal Models}

\author[1]{Jo\~ao P.\ Arroyo}
\author[1]{Jo\~ao G.\ Rodrigues}
\author[1]{Daniel Lawand} 
\author[1]{Denis D.\ Mau\'a}
\author[2]{\hspace*{2mm}Junkyu Lee}
\author[2]{\hspace*{2mm}Radu Marinescu}
\author[3]{Alex Gray}
\author[4]{Eduardo R.\ Laurentino}
\author[1]{Fabio G.\ Cozman}
\affil{\vspace*{-3.48ex}\mbox{~}}
\affil[1]{
    Universidade de S\~ao Paulo, S\~ao Paulo, Brazil
}
\affil[2]{%
    IBM Research -- J.\ L.: Yorktown Heights, USA; R.\ M.: Ireland
}
\affil[3]{%
    Centaur Institute, USA
}
\affil[4]{%
    Instituto de Ci\^encia e Tecnologia Ita\'u\\ S\~ao Paulo, Brazil
}

\begin{document}

\maketitle

\begin{abstract}
We investigate partially identifiable queries in a class of causal models.
We focus on acyclic Structural Causal Models that are quasi-Markovian (that is, each endogenous variable is connected with at most one exogenous confounder). We look into scenarios where endogenous variables are observed (and a distribution over them is known), while exogenous variables are not fully specified. 
This leads to a representation that is in essence a Bayesian network where the distribution of root variables is not uniquely determined. 
In such circumstances, it may not be possible to precisely compute a probability value of interest.
We thus study the computation of tight probability bounds, a problem that has been solved by  multilinear programming in general, and by linear programming when a single confounded component is intervened upon. 
We present a new algorithm to simplify the construction of such programs by exploiting input probabilities over endogenous variables.
For scenarios with a single intervention, we apply column generation to compute a probability bound through a sequence of auxiliary linear integer programs, thus showing that a representation with polynomial cardinality for exogenous variables is possible. Experiments show column generation techniques to be superior to existing methods.
\end{abstract}

\section{Introduction}\label{section:Introduction}

Structural Causal Models (SCMs) offer a representation where some variables are associated with deterministic mechanisms while other  variables are associated with marginal probabilities.
We refer to the former variables as endogenous ones, and to the latter variables as exogenous ones.

It is often the case that observational data determines the probability distribution of endogenous variables, but not the distribution of exogenous variables. In fact, the structure of exogenous variables may not be specified, and all that one may know are bounds on the cardinalities of those variables. 
In such a setting, the causal diagram that captures the connections between variables can be viewed as a Bayesian network where the (marginal) distribution of root variables is not specified \citep{Zaffalon2020}. This is the sort of representation we explore in this paper.

Given such a causal model, one may be interested in some probability value (or some function of probability values) under interventions. When such a query leads to a single precise number, we say the query is identifiable. Pearl's do-calculus  can be used to determine when a query is identifiable  \citep{causality}. 

When identifiability fails, one can still bound probability values given a distribution over the observed variables.
Hopefully, one may then produce probability intervals that are sufficiently informative to make decisions
--- when the observable variables are discrete, this can be achieved without additional assumptions on the unknown mechanisms at play \citep{balke-pearl97}.  

In this paper we focus on the computation of tight probabilistic bounds when identifiability fails; that is, we focus on the partially identifiable setting. We abuse language by using ``SCM'' to also refer to models where the marginal distribution over exogenous variables is not known precisely, but rather imprecisely induced by a distribution over endogenous variables.

We restrict interest to {\em quasi-Markovian} SCMs; that is, to SCMs where each endogenous variable has at most one exogenous parent. This family of SCMs is quite expressive; for instance, it contains Balke and Pearl's imperfect compliance example and many of their extensions in the literature. Moreover, quasi-Markovian SCMs can be used to approximate non-quasi-Markovian ones \citep{zhang-tian-bareinboim2022}.

There have been several relevant proposals to cast partially identifiable queries with quasi-Markovian SCMs as nonlinear programs that, in some cases, reduce to linear programs.
For example, \citet{balke-pearl} showed that linear programming can bound the causal effect of an intervention in the so-called instrumental variable model. 
\citet{tian-pearl} then showed how to write a linear program to compute the probability of necessity and sufficiency in two-variable binary models. 
\citet{sachs} extended \citet{balke-pearl}'s models to a larger class,  for which they showed that causal effect inferences can be cast as linear programs; the size of these linear programs has been studied, and in many cases reduced, by \citet{Shridharan2024}.
\citet{zhang-tian-bareinboim2022} described techniques that bound the cardinality of non-observed variables and that can be used to approximate bounds on causal effects through linear programming. 
\citet{Duarte2024} instead focused on the general multilinear programs that produce anytime probability bounds (in the sense that the algorithm can be stopped at any given time and still yield  approximate bounds). 
\citet{zaffalon2023} translated the computation of causal inferences to inference in credal networks, and looked at approximate solutions based on sequences of linear programs and on parametric learning (a credal network is, in essence, a Bayesian network where conditional probability distributions are only known to belong to specified sets of distributions  \citep{Cozman2000}).

Recently, \citet{shridharan23icml} derived a significant result for quasi-Markovian SCMs.
In short, for those SCMs,  tight bounds on causal effects can be computed by multilinear programs whose degree is restricted to the number of intervened confounded components --- and hence to linear programming when a single confounded component is intervened upon!
%
It is within the context of Shridharan and Iyengar's work that we make our contributions.

We first present a new proof for the main result by \citet{shridharan23icml}, a proof that hopefully shows this clever result to be in essence a simple one from a mathematical point of view (Section \ref{section:ShortProof}). We then present a new algorithm that shows how to exploit information in an input  distribution over endogenous variables to simplify the construction of such multilinear/linear programs (Section~\ref{section:Algorithm}). Our new algorithm exploits Pearl's do-calculus to build up a simplified objective function.

One practical challenge is that bounds on the cardinality of exogenous variables may be very large, thus leading to large multilinear/linear programs.
To avoid this explosion,  we introduce column generation techniques; that is, we show how to build a sequence of basis changes that lead to the desired bound (Section \ref{sec:columngen}). Each change of basis uses a common master linear program and an auxiliary program that, despite the presence of polynomial terms in its specification, can be reduced to a linear integer program. 
We also show how to build a {\em single} linear integer program, 
that directly produces desired bounds. 
We present empirical evidence that our proposed techniques are   superior to existing approaches.

\begin{figure*}
    \centering
    \begin{tikzpicture}[>=latex]
      \tikzstyle{endo} = [draw,circle,fill=black,minimum size=4pt,inner sep=0.0em]
      \tikzstyle{exo} = [draw,circle,fill=white,minimum size=4pt,inner sep=0.0em]
      \node[exo,label=left:$U_1$] (U1) at (0.5,.75) {};
      \node[exo,label=right:$U_2$] (U2) at (2.5,.75) {};
      \node[endo,label=below:$X$] (A) at (0,0) {};
      \node[endo,label=below:$W$] (B) at (1,0) {};
      \node[endo,label=below:$Z$] (C) at (2,0) {};
      \node[endo,label=below:$Y$] (D) at (3,0) {};
      \foreach \x/\y in {U1/A,U1/B,A/B,B/C,U2/C,U2/D,C/D} {
        \draw[->] (\x) edge (\y);
      };
      \node at (1.5,0.75) {(a)};
    \end{tikzpicture} 
    \hspace*{5mm}
    \begin{tikzpicture}[>=latex]
      \tikzstyle{endo} = [draw,circle,fill=black,minimum size=4pt,inner sep=0.0em]
      \tikzstyle{exo} = [draw,circle,fill=white,minimum size=4pt,inner sep=0.0em]
      \node[endo,label=below:$X$] (A) at (0,0) {};
      \node[endo,label=below:$W$] (B) at (1,0) {};
      \node[endo,label=below:$Z$] (C) at (2,0) {};
      \node[endo,label=below:$Y$] (D) at (3,0) {};
      \foreach \x/\y in {A/B,B/C,C/D} {
        \draw[->] (\x) edge (\y);
      };
      \draw[->] (B) to [bend left] (D);
      \node at (1.5,0.75) {(b)};
    \end{tikzpicture} 
    \hspace*{5mm}
    \begin{tikzpicture}[>=latex]
      \tikzstyle{endo} = [draw,circle,fill=black,minimum size=4pt,inner sep=0.0em]
      \tikzstyle{exo} = [draw,circle,fill=white,minimum size=4pt,inner sep=0.0em]
      \node[exo,label=left:$U_1$] (U1) at (0.5,.75) {}; 
      \node[endo,label=below:$X$] (A) at (0,0) {};
      \node[endo,label=below:$W$] (B) at (1,0) {};
      \node[endo,label=below:$Z$] (C) at (2,0) {};
      \node[endo,label=below:$Y$] (D) at (3,0) {};
      \foreach \x/\y in {U1/A,U1/B,A/B,B/C,C/D} {
        \draw[->] (\x) edge (\y);
      };
      \draw[->] (B) to [bend left] (D);
      \node at (1.5,0.75) {(c)};
    \end{tikzpicture} 
        \hspace*{5mm}
    \begin{tikzpicture}[>=latex]
      \tikzstyle{endo} = [draw,circle,fill=black,minimum size=4pt,inner sep=0.0em]
      \tikzstyle{exo} = [draw,circle,fill=white,minimum size=4pt,inner sep=0.0em]
      \node[exo,label=left:$U_1$] (U1) at (0.5,.75) {}; 
      \node[endo,label=below:$X$] (A) at (0,0) {};
      \node[endo,label=below:$W$] (B) at (1,0) {};
      \node[endo,label=below:$Z$] (C) at (2,0) {};
      \node[endo,label=below:$Y$] (D) at (3,0) {};
      \foreach \x/\y in {U1/B,A/B,B/C,C/D} {
        \draw[->] (\x) edge (\y);
      };
      \draw[->] (B) to [bend left] (D);
      \node at (1.5,0.75) {(d)};
    \end{tikzpicture} 
    \vspace*{-1ex}
   \caption{(a) An example proposed by \cite{sachs}. 
   (b) The factorization of the (marginal) distribution for endogenous variables.
   (c) A semi-marginal graph that marginalizes $U_2$.
   (d) The intervened semi-marginal graph for $\Do{X=x}$.}
    \label{fig:examples}
\end{figure*}
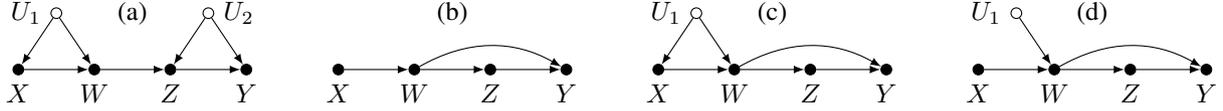

\section{Background}\label{section:Background}

We write random variables using capital letters (e.g., $X$) and sets of random variables using boldface (e.g., $\mathbf{X}$). The support of $X$  is denoted as $\val({X})$, and  $\val({\mathbf{X}})$ is the direct product of the support of each variable in the set. 
A probability value is denoted by  $\Pr(\mathbf{X}=\mathbf{x})$, or 
$\Pr(\mathbf{x})$ when appropriate.
A distribution is denoted by  $\Pr(\mathbf{X})$.
We consider graphs whose nodes are random variables, and refer indiscriminately to nodes and random variables.
We   write $\pa{X}$ to denote the parents of $X$ in a graph $\graph{G}$.
We write $\graph{G}_{\underline{\mathbf{X}}}$ to denote the graph obtained from $\mathcal{G}$ by removing all edges leaving nodes in the node set $\mathbf{X}$,
and  $\graph{G}_{\overline{\mathbf{X}}}$ to denote the graph obtained by removing edges entering nodes in $\mathbf{X}$.

    A \emph{Structural Causal Model} (SCM) is a tuple $(\mathcal{G},\mathbf{V},\mathbf{U},\mathcal{F},\Pr(\mathbf{U}))$ where $\graph{G}$ is an directed graph whose node set is $\mathbf{V} \cup \mathbf{U}$, where $\mathbf{V}$ are the inner nodes, called \emph{endogenous}, and $\mathbf{U}$ are the root nodes, called \emph{exogenous}; $\mathcal{F}$ is a set of functions $f_V: \val(\pag{V }) \rightarrow \val(V )$ called \emph{mechanisms}, one for each node $V$ in $\mathcal{G}$; and $\Pr(\mathbf{U})$ is a probability distribution over the exogenous random variables $\mathbf{U}$ \citep{scm,halpern}.

We restrict ourselves here to acyclic graphs. 
For such models, we have for any endogenous variable $V \in \mathbf{V}$ that
\[
\Pr(V=v|\pa{V}=\pi) = \bracket{f_V(\pi)=v},
\]
where $\bracket{\theta}$ denotes the Iverson bracket (i.e., it is 1 if statement $\theta$ holds, and 0 otherwise). 
We assume  exogenous nodes are independent, hence $\Pr(\mathbf{u})=\prod_{U \in \mathbf{U}} \Pr(U=u)$.
Consequently,  
\begin{equation}
\label{equation:Joint}
\Pr( \mathbf{v}, \mathbf{u}) = 
\prod_V \bracket{f_V(\pi)=v} \prod_U \Pr(U=u),
\end{equation}
for values of  $v$, $\pi$ and $u$ that are consistent with $\mathbf{v}$ and $\mathbf{u}$.
The latter expression is multilinear on the probabilities of exogenous variables. Any marginal probability is therefore a multilinear expression of $\{ \Pr(U=u): u \in \val(U), U \in \mathbf{U}\}$.

A \emph{confounded component} (for short, c-component) of a directed graph $\graph{G}$ is a set of endogenous nodes in a maximal connected component of the undirected version of  graph $\graph{G}_{\underline{\mathbf{V}}}$ (i.e., the graph obtained by removing endogenous-to-endogenous edges) \citep{tian}.
 
    An SCM is \emph{quasi-Markovian} if every
    endogenous variable has at most one exogenous variable as parent.
    If in addition every exogenous variable has exactly one child, then the model is said to be Markovian.
Figure \ref{fig:examples} (a) shows the graph of a quasi-Markovian SCM (that is not Markovian), taken from a rather simple example by \citep{sachs}. The model has two c-components: $\{X,W\}$ and $\{Z,Y\}$.

Consider a c-component $\mathbf{C}$ of a quasi-Markovian SCM, and let $U$ be the (single)  exogenous parent of   nodes in $\mathbf{C}$, and $\mathbf{W}_{\mathbf{C}}$ denote the union of the variables in $\mathbf{C}$ and all of their endogenous parents. 
A key result by \cite{tian} is that:
\begin{equation}
\label{equation:CFactors}
 \Pr(\mathbf{V}=\mathbf{v}) = \prod_{\mathbf{C}} Q_{\mathbf{C}}(\mathbf{w}_\mathbf{C}),
\end{equation}
where
$\mathbf{w}_\mathbf{C}$ is the configuration of $\mathbf{W}_{\mathbf{C}}$ that is consistent with $\mathbf{v}$, and
\begin{equation}
\label{equation:CFactor}
Q_{\mathbf{C}}(\mathbf{W}_{\mathbf{C}}) = \sum_u \Pr(u) \prod_{V \in \mathbf{C}} \bracket{f_V(\pa{V})=V}. 
\end{equation}
For $V \in \mathbf{C}$, let $\mathbf{W}_V$ denote the variables that are topologically smaller than $V$ in $\mathbf{W}_\mathbf{C}$. 
\citet{tian} also showed that:
\begin{equation}
\label{equation:TianFactor}
Q_{\mathbf{C}}(\mathbf{W}_{\mathbf{C}}) = \prod_{V \in \mathbf{C}}  \Pr(V|\mathbf{W}_{V}).
\end{equation}
As c-components form a partition of $\mathbf{V}$, we   have that: 
\begin{equation}
    \label{equation:TianFactorization}
 \Pr(\mathbf{V}) = \prod_{V \in \mathbf{V}} \Pr(V|\mathbf{W}_{V}) \, .
\end{equation}
For example, for the graph  in Figure~\ref{fig:examples} (a), Equation~\eqref{equation:TianFactorization} implies 
$\Pr(W,X,Y,Z) = \Pr(X) \Pr(W|X)$ $\Pr(Z|W)\Pr(Y|W,Z)$
(as captured by Figure~\ref{fig:examples} (b)).

As \cite{zaffalon2023} noted, Equations  (\ref{equation:CFactor}) and (\ref{equation:TianFactor}) lead to necessary and sufficient linear constraints over $\Pr(U)$:
\begin{equation} \label{eq:linearconstraint}
     \prod_{~~~V \in \mathbf{C}} \!\!\!\!\Pr(V|\mathbf{W}_V)     
     = \sum_u \Pr(u) \!\!\!\!\prod_{~~~V \in \mathbf{C}} \!\!\!\bracket{f_V(\pa{V})\!=\!V}.
\end{equation}
Note that there is one constraint per configuration of $\mathbf{W}_\mathbf{C}$.
 
When endogenous random variables are categorical, one can always extend a causal graph into a partially specified SCM, that is, an SCM without a fixed exogenous distribution $\Pr(\mathbf{U})$.
That process is known as \emph{canonicalization} \citep{zhang-tian-bareinboim2022}, and  essentially, consists of enumerating all possible mechanisms via values of the exogenous variables.
For quasi-Markovian graphs, canonicalization reduces each exogenous variable  $U$ to a categorical random variable whose state space has cardinality $\prod_{V \in \mathbf{C}}|\val(V)|^{|\val(\pa{V})|}$, where $\mathbf{C}$ is the corresponding c-component. 
Each value $u \in \val(U)$ specifies a mechanism $f_V: \pag{V} \rightarrow V$ for each $V$ in the c-component.
Thus, we assume without loss of generality   that every exogenous variable is categorical.

A simple \emph{intervention}  $\doX$ modifies an SCM by substituting $f_X$ with $\bracket{X=x}$, where $x$ is the corresponding value in $\mathbf{x}$, for every $X \in \mathbf{X}$.
Graphically, interventions are represented by means of surgery of $\mathcal{G}$, producing $\mathcal{G}_{\overline{\mathbf{X}}}$. 

An intervention induces a new (post-intervention) distribution over any variable set $\mathbf{Y}$, denoted as $\Pr(\mathbf{Y}|\dox)$.
The goal of causal inference is to estimate expressions involving such probabilities using the constraints shared by the non-intervened and intervened SCMs.
We are interested in the calculation of population-level causal effects that can be written as linear combinations of post-intervention probabilities, such as the average treatment effect (ATE) and the conditional average treatment effect (CATE).
For the sake of exposition, we concentrate on simple inferences such as $\Pr(\mathbf{Y}|\dox)$; however, the same algorithms and results apply for linear combinations of such probabilities.

\section{Computing Probability Bounds}
\label{section:ShortProof}

Suppose we have a partially specified quasi-Markovian SCM $(\mathcal{G},\mathbf{V},\mathbf{U},\mathcal{F})$  and an {\em input} distribution $\widehat{\Pr}(\mathbf{V})$. 
In the remainder of this paper we assume endogenous variables are binary; this substantially simplifies the presentation by reducing the number of necessary indexes (extension to categorical variables should be clear). 

We are interested in computing $\Pr(\mathbf{Y}=\mathbf{y}|\doX)$, abbreviated $\Pr(\mathbf{y}|\dox)$.
We do not assume identifiability; that is, the input distribution may not be sufficient to constrain $\Pr(\mathbf{y}|\dox)$ to a point value; for this reason, we wish to compute the \emph{lower probability} defined as
\[
\underline{\Pr}(\mathbf{y}|\dox) := \inf \Pr(\mathbf{y}|\dox),
\]
where the infimum is taken over the set of all extensions of the given partially specified SCM whose induced joint distributions satisfies the set of constraints given by Equation~\eqref{eq:linearconstraint}, with 
\[
\Pr(V|\mathbf{W}_V)=\widehat{\Pr}(V|\mathbf{W}_V).
\]
\cite{zaffalon2023} have shown this optimization problem to lead to tight bounds. 
Note that the fact that the feasible region is a closed polytope allows us to replace 
infimum by minimum in the expression above, so
\[
\underline{\Pr}(\mathbf{y}|\dox)   = \min \Pr(\mathbf{y}|\dox).
\]
One might as well be interested in the {\em upper probability}, by taking maximum instead of minimum; the necessary changes should be obvious.


As shown by \citet{shridharan23icml}, the multilinear expression for $\Pr(\mathbf{y}|\dox)$ need only   involve probabilities $\Pr(U=u)$ for exogenous variables connected with c-components intervened by $\doX$.\footnote{This assumes that the optimization is feasible.} Thus, the degree of the multilinear expression is equal to the number of intervened c-components. This result was  proved by them by explicitly developing the relevant factors in the multilinear expression; we now present a shorter argument directly based on Tian's factorization given by Expression~\eqref{equation:CFactors}.

To start, let $\mathbf{Z}=\mathbf{V}\!\setminus\!\{\mathbf{X},\mathbf{Y}\}$. 
Thus $\Pr(\mathbf{y}|\dox)$ is equal to
\[ 
\sum_{\mathbf{z}} \Pr(\mathbf{y},\mathbf{z}, \mathbf{x}|\dox) =
\sum_{\mathbf{z}} \prod_{\mathbf{C}} Q_\mathbf{C}^{\dox}(\mathbf{w}_\mathbf{C}) ,
\]
where the latter equality follows from Expression \eqref{equation:CFactors} with each $\mathbf{w}_\mathbf{C}$ being consistent with $\mathbf{y}$, $\mathbf{z}$ and $\mathbf{x}$.
To proceed, let $\mathcal{C}_0$ denote the collection of c-components of the non-intervened model that \emph{do not} contain intervened variables.
Note that a c-component $\mathbf{C}\in\mathcal{C}_0$ does {\em not} change under the intervention, as an intervention $\doX$ modifies only the mechanisms related to $X \in \mathbf{X}$.
As a consequence, in the expression above, each term $Q_\mathbf{C}^{\dox}$ that refers to a  c-component  $\mathbf{C}\in\mathcal{C}_0$ equals the analogous term $Q_\mathbf{C}$ in the factorization of the non-intervened SCM (as both are given by identical instances of Expression (\ref{equation:CFactor})). 
Moreover, by Equation \eqref{equation:TianFactor} $Q_\mathbf{C}$ can be written as a product of empirical probabilities $\widehat{\Pr}(v|\mathbf{w}_V)$ for each $V \in \mathbf{C}$.
Similarly, for each $\mathbf{C}$ not in $\mathcal{C}_0$ the term $Q_\mathbf{C}^{\dox}$ in the expression above differs from the analogous term $Q_\mathbf{C}$ only with respect to the mechanisms for the intervened variables, which become  $\bracket{X=x}$.
Connecting all observations above, we get:
\begin{multline} \label{equation:MixedFactorization}
\Pr(\mathbf{y}|\dox) = 
\sum_\mathbf{z} 
\prod_{\mathbf{C} \in \mathcal{C}_0} 
  \;\; \prod_{V \in \mathbf{C}}  \widehat{\Pr}(v|\mathbf{w}_{V}) 
  \times \\ 
\prod_{\mathbf{C} \not\in \mathcal{C}_0} 
  \sum_u \Pr(u) \prod_{V \in \mathbf{C}\backslash\mathbf{X}}  
  \bracket{f_V(\pi)=v}
 ,
\end{multline}
where all assignments inside the outer summation must agree with the assignments for $\mathbf{Y}=\mathbf{y}$, $\mathbf{Z}=\mathbf{z}$, and $\mathbf{X}=\mathbf{x}$.
Now define $\mathbf{U}^*$ to be the set of exogenous variables that are parents of intervened c-components.
Then we have that 
\[
\underline{\Pr}(\mathbf{y}|\dox) = \min_{\Pr(U):U\in\mathbf{U}^*} \Pr(\mathbf{y}|\dox),
\]
where $\Pr(\mathbf{y}|\dox)$ is given by 
Expression~(\ref{equation:MixedFactorization}), 
subject to the set of constraints given by Equation~\eqref{eq:linearconstraint}.
As that defines a multilinear program, we  conclude our proof  of \citet{shridharan23icml}'s main result.

It is actually possible to convey this latter result in a rather visual way by using graphs.
A simple example should clarify the idea.
Consider an SCM with graph as in Figure~\ref{fig:examples} (a), and an intervention $\Do{X=x}$.
We may marginalize all exogenous variables (Figure \ref{fig:examples} (b)), or just $U_2$ (Figure \ref{fig:examples} (c)); the latter produces what we call a {\em semi-marginal graph}.
Equation \eqref{equation:MixedFactorization} denotes the factorization given by Equation~\eqref{equation:Joint} in the graph in Figure~\ref{fig:examples} (d).
In the latter, the exogenous variable $U_1$, associated with the intervened c-component $\{W,X\}$ of the original graph, is kept while the exogenous $U_2$ in the non-intervened c-component $\{Z,Y\}$ is discarded  ---
note the edge $W \rightarrow Y$ is  included to account for the missing exogenous variable. 

That is, we can take the directed graph, create a topological order for the variables, and 
for each c-component $\mathbf{C}$ that is not intervened, 
\begin{itemize}
\item remove the corresponding exogenous variable, and 
\item connect each variable $V$ in $\mathbf{C}$ with the variables in $\mathbf{W}_V$
(recall that $\mathbf{W}_V$ contains the variables that are topologically smaller than $V$ 
among the variables in $\mathbf{C}$ and the variables that are parents of variables in $\mathbf{C}$);
\end{itemize}
 and then remove edges corresponding to the interventions as usual.
We refer to this latter type graph as the {\em intervened semi-marginal} graph,  that is, the graph obtained by marginalizing exogenous variables connected with non-intervened c-components and performing surgery on the intervened variables.
Additional intervened semi-marginal graphs can be found in the Supplementary Material.

\section{Exploiting Input Distributions}
\label{section:Algorithm}

Suppose we have a quasi-Markovian causal graph $\mathcal{G}$, an input distribution $\widehat{\Pr}(\mathbf{V})$, target variables $\mathbf{Y}$ and a \emph{single} intervention $\Do{X=x}$. 
Then \cite{shridharan23icml}'s result show us that $\Pr(\mathbf{y}|\Do{x})$ can be cast as a linear program whose objective function, given by Equation~\eqref{equation:MixedFactorization},  contains $\val(U \cup \mathbf{Z})$ terms.
We can instead simplify the expression for instance by running symbolic variable elimination  \citep{pgm} with the factors defined by the semi-marginal graph.
Doing so leads to an expression whose number of terms is given by $\val(U)$ times an exponential in the graph's treewidth (which measures the connectivity of the causal graph).

However, we can do better. As we now show, we may require only a smaller set of factors, and less exogenous variables, by taking a different route.

As a preliminary point, note that any node that is not in the ancestral graph of $\mathbf{Y}$ can be discarded, as it will not affect the result of the causal inference. 
This can   be seen from the intervened semi-marginal graph, which in that case will have such a  node d-separated from $\mathbf{Y}$ or barren.
That holds even for intervened $X \in \mathbf{X}$; we thus assume that each $X$ is an ancestor of some $Y \in \mathbf{Y}$ in the following.

Now consider simplifications that exploit the input distribution. To do so, we take the input distribution as an oracle that can yield any conditional   distribution   $\widehat{\Pr}(\mathbf{R}|\mathbf{S})$, for any $\mathbf{R},\mathbf{S} \subseteq \mathbf{V}$. In practice, efficient data structures can be used to obtain such probabilities from e.g.\ a dataset (note: ``local'' probabilities are needed anyway to handle Expression \eqref{equation:MixedFactorization}).

We present an algorithm that produces a sequence of expressions for auxiliary values $\Pr(\mathbf{Y}_t|\Do{x})$;
the objective function can be built at the end by collecting such expressions, leading to an expression that potentially is exponentially more succinct that the one given by Expression \eqref{equation:MixedFactorization}. 

Starting with the input target set, $\mathbf{Y}_0=\mathbf{Y}$, at each step $t=0,1,\dotsc,$ the algorithm selects a variable $Y_t \in \mathbf{Y}_t$ that has no descendants in $\mathbf{Y}_t$; that is, the algorithm selects variables $Y_0 < Y_1 < \dotsb$ in reverse topological order. It then defines a next target set $\mathbf{Y}_{t+1}$ and produces a symbolic expression relating $\Pr(\mathbf{Y}_t|\Do{x})$ and $\Pr(\mathbf{Y}_{t+1}|\Do{x})$, according to the following cases.

\begin{enumerate}[label=\textbf{C\arabic*}:,ref={C\arabic*}]
\item \label{cond1a} {\bf $Y_t$ is in $\mathbf{U}$ (hence $Y$ is not a descendant of $X$).} \\
Then build $\mathbf{Y}_{t+1} = \mathbf{Y}_t \setminus \{Y_t\}$ and output
\[
    \Pr(\mathbf{y}_t|\Do{x}) =  \Pr(\mathbf{y}_{t+1}|\Do{x}) \Pr(Y_t = u) 
\]
for each assignment $\mathbf{y}_t$ of $\mathbf{Y}_t$, where $u$ agrees with $\mathbf{y}_t$. If $\mathbf{Y}_{t+1}=\emptyset$ then set $\Pr(\mathbf{y}_{t+1}|\Do{x})=1$.
\item \label{cond1b} 
{\bf $Y_t \in \mathbf{V}$ is not a descendant of $X$.} \\ 
Then build $\mathbf{Y}_{t+1} = \mathbf{Y}_t \setminus \{Y_t\}$ and output
\[
    \Pr(\mathbf{y}_t|\Do{x}) =  \Pr(\mathbf{y}_{t+1}|\Do{x})\widehat{\Pr}(y|\mathbf{y}_{t+1}) 
\]
for each assignment $\mathbf{y}_t$ of $\mathbf{Y}_t$.
\item \label{cond2} {\bf $Y_t$ is a descendant of $X$ and they  are in the same c-component.} \\
Then define $\mathbf{Z}_t= \pa{Y_t} \setminus (\mathbf{Y}_t \cup \{X\})$, and $\mathbf{Y}_{t+1}=(\mathbf{Y}_t \cup \mathbf{Z}_t) \setminus \{Y_t\}$.
For each assignment $\mathbf{y}_t$ of $\mathbf{Y}_t$, output
\[
\Pr(\mathbf{y}_t | \Do{x}) = \sum_{\mathbf{z}_t} \bracket{f_{Y_t}(\pi)=y_t} \Pr(\mathbf{y}_{t+1}|\Do{x}) ,
\]
where $\pi$ is the assignment of the parents of $Y_t$ that agrees with $\mathbf{z}_t$, $\mathbf{y}_t$ and $x$, and $y_t$ agrees with $\mathbf{y}_t$.
\item \label{cond3} {\bf $Y_t$ is a descendant of $X$ but they are not in the same c-component.} \\
Define $U_t = \pa{X}$. 
Find a set of endogenous variables $\mathbf{W}_t$ such that (i) $Y_t$ and $X$ are d-separated by $\mathbf{S}_t$ in $\mathcal{G}_{\underline{X}}$ (the graph where edges leaving $X$ are removed), and (ii) $Y_t$ and $U_t$ are d-separated by $\mathbf{S}_t \cup \{ X \}$ in $\graph{G}$, where $\mathbf{S}_t = (\mathbf{W}_t \cup \mathbf{Y}_t) \setminus \{ Y_t,U_t \}$.
Let $\mathbf{Z}_t = \mathbf{W}_t \setminus \mathbf{Y}_t$.
Build  $\mathbf{Y}_{t+1} = (\mathbf{Y}_t \cup \mathbf{Z}_t) \setminus \{Y_t\}$ and output
\[
\Pr(\mathbf{y}_t | \Do{x}) = \sum_{\mathbf{z}_t} \widehat{\Pr}(y_t|x,\mathbf{s}_t) \Pr(\mathbf{y}_{t+1} | \Do{x}) \, ,
\]
where the assignments on the right hand side agree with the assignments on the left hand side.
\end{enumerate}

Each one of these cases produces a new expression; as noted previously,
these expressions can be collected 
to generate a single linear expression for $\Pr(\mathbf{y}|\Do{x})$ that contains   optimization variables $\Pr(u)$ (for each $u$).

Before we prove the algorithm correct, consider a few examples that convey its behavior.

\begin{example}
Consider the graph in Figure~\ref{fig:examples} (a), an intervention $\Do{x}$ and target $\mathbf{Y}_0 = \{Y\}$. 
At step $t=0$, the algorithm selects $Y$.
To apply \ref{cond3}, we find a set $\mathbf{W}_0$ that d-separates $Y$ and $X$ in $\graph{G}_{\underline{X}}$, and also d-separates $Y$ and $U_1$ in $\mathcal{G}$. The only set satisfying such conditions  is $\mathbf{W}_0=\{W\}$.
The algorithm produces
\[
 \Pr(y|\Do{x}) = \textstyle\sum_w \widehat{\Pr}(y|x,w)\Pr(w|\Do{x}) .
\]
Then the algorithm moves to $\mathbf{Y}_1=\{W\}$, selects $W$, which satisfies \ref{cond2}, and produces
\[
  \Pr(w|\Do{x}) = \textstyle\sum_{u_1} \bracket{f_W(x,u_1)=w}\Pr(u_1|\Do{x}).
\]
Last, the algorithm takes $\mathbf{Y}_2=\{U_1\}$, satisfying \ref{cond1a}, and produces: $\Pr(u_1|\Do{x})=\Pr(u_1)$.
Collecting all expressions, we get a linear  objective function:
\[
  \textstyle \sum_w \widehat{\Pr}(y|x,w)\sum_{u_1} \bracket{f_W(x,u_1)=w}\Pr(u_1) \, .  \qquad \Box
\]
\end{example}


\begin{figure}[t]
    \centering
    \begin{tikzpicture}[>=latex]
      \tikzstyle{endo} = [draw,circle,fill=black,minimum size=4pt,inner sep=0.0em]
      \tikzstyle{exo} = [draw,circle,fill=white,minimum size=4pt,inner sep=0.0em]
      \node[exo,label=right:$U_1$] (U1) at (1,0) {};
      \node[exo,label=left:$U_2$] (U2) at (0,0) {};
      \node[endo,label=below:$X$] (A) at (0,-0.75) {};
      \node[endo,label=below:$Y$] (B) at (2,-0.75) {};
      \node[endo,label=below:$R$] (C) at (-1,-.75) {};
      \node[endo,label=below:$Z$] (D) at (1,-.75) {};
      \foreach \x/\y in {A/D,D/B,A/C,C/D,U1/A,U1/B,U2/C,U2/D} {
        \draw[->] (\x) edge (\y);
      };
      \draw[->] (C) to [out=20,in=160] (D);
      \draw[->] (A) to [out=20,in=160] (B);
    \end{tikzpicture}
    \hspace*{10mm}
    \begin{tikzpicture}[>=latex]
      \tikzstyle{endo} = [draw,circle,fill=black,minimum size=4pt,inner sep=0.0em]
      \tikzstyle{exo} = [draw,circle,fill=white,minimum size=4pt,inner sep=0.0em]
      \node[exo,label=left:$U_1$] (U1) at (0,.75) {};
      \node[exo,label=right:$U_2$] (U2) at (1,.75) {}; 
      \node[endo,label=below:$Z$] (Z) at (0,0) {};
      \node[endo,label=below:$Y$] (Y) at (2,0) {};
      \node[endo,label=below:$X$] (X) at (-1,0) {};
      \node[endo,label=below:$W$] (W) at (1,0) {};
      \foreach \x/\y in {X/Z,Z/W,W/Y,U1/X,U1/W,U2/Z,U2/Y} {
        \draw[->] (\x) edge (\y);
      };
      \draw[->] (X) to [out=20,in=160] (W);
    \end{tikzpicture}
    \caption{Graphs for quasi-Markovian SCMs, used in examples. 
    None of them are handled by techniques by \cite{sachs}.}
    \label{fig:nonlinear2}
\end{figure}
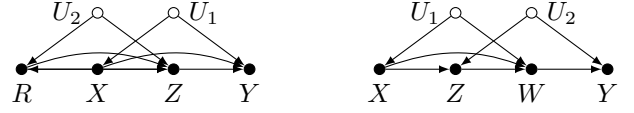

\begin{example}
Consider the graph in Figure~\ref{fig:nonlinear2} (left). Given intervention $\Do{x}$ and target $\mathbf{Y}_0=\{Y\}$, the algorithm selects variables in the ordering $Y$, $Z$ and $U_1$, creates   sets $\mathbf{Y}_1=\{Z,U_1\}$, $\mathbf{Y}_2=\{U_1\}$, and outputs, respectively:
    \begin{align*}
        \Pr(y|\Do{x}) &= \sum_{z,u_1: f_Y(z,x,u_1)=y} \Pr(z,u_1|\Do{x}) .
        & \text{[\ref{cond2}]}
        \\
        \Pr(z,u_1|\Do{x}) &= \widehat{\Pr}(z|x)\Pr(u_1|\Do{x}) .
        & \text{[\ref{cond3}]}
        \\
        \Pr(u_1|\Do{x}) &= \Pr(u_1) .
        & \text{[\ref{cond1a}]}
    \end{align*}
Collecting all expressions, we get a linear objective function
 {\em that notably does not mention $R$}:
\[
 \textstyle \sum_{z,u_1}\bracket{f_Y(z,x,u_1)=y}\widehat{\Pr}(z|x)\Pr(u_1) \, . \qquad \Box
\]
\end{example}

\begin{example}
Consider a graph as in Figure~\ref{fig:nonlinear2} (left), but with the edge $R \rightarrow Z$ replaced by a sequence $R \rightarrow R_1 \rightarrow \dotsb \rightarrow R_n \rightarrow Z$, where each $R_i$ is also  connected to $U_1$. Then Expression~\eqref{equation:MixedFactorization} generates an expression with over $|\val(U_1)|2^n$ terms, whereas our algorithm generates the same expression as in the previous example. And if we replace each $R_i$ by a subgraph with high treewidth, the same example shows that our algorithm can produce expressions that are exponentially smaller then by running symbolic variable elimination on the intervened semi-marginal graph. $\Box$
\end{example}

Now consider correctness:

\begin{theorem} \label{thm:alg-is-sound}
 The previous algorithm generates a linear program that, when optimized subject to constraints given by Equation~\eqref{eq:linearconstraint},  computes tight bounds for $\Pr(\mathbf{y}|\Do{x})$.
\end{theorem}

The proof requires the following lemma.

\begin{lemma} \label{lem:dsep-for-case4}
    Suppose a quasi-Markovian graph $\graph{G}$ with endogenous nodes $X$, $Y$  and $\mathbf{Z}$ is such that: 
    (i) $X$ and $Y$ have no common exogenous parent;
    (ii) $Y$ is a descendant of $X$;
    (iii) there are no descendants of $Y$ in $\mathbf{Z}$. 
    Then there is a set of endogenous variables $\mathbf{W}$ that are ancestors of $\mathbf{Z} \cup \{Y\}$ and such that $X$ and $Y$ are d-separated by $\mathbf{W} \cup \mathbf{Z}$ in $\graph{G}_{\underline{X}}$. In addition, $\mathbf{W} \cup \mathbf{Z} \cup \{X\}$ also d-separates $Y$ and $U$ in $\graph{G}$, where $U=\pa{X} \cap \mathbf{U}$.
\end{lemma}

\begin{proof}[Proof of Lemma~\ref{lem:dsep-for-case4}]
    Consider the moral graph $\graph{M}$ of the ancestors of $\mathbf{Z}$ and $Y$ in $\graph{G}_{\underline{X}}$, which also include the ancestors of $X$ by Assumption~(ii). Then, $X$ and $Y$ are d-separated by some superset of $\mathbf{Z}$ in $\graph{G}_{\underline{X}}$ iff there is no undirected path in $\graph{M}$ containing no endogenous nodes (other than $X$ and $Y$) \citep{pgm}. First note that $\graph{M}$ has no edge $X$--$Y$, since $X$ has no outgoing edges in $\graph{G}_{\underline{X}}$, $Y$ is descendant of $X$ by Assumption~(ii) and by Assumption~(iii) $X$ and $Y$ have no common child in $\mathcal{M}$. Thus consider a path between $X$ and $Y$ containing $U$. By Assumption~(i) and because $\graph{G}$ is quasi-Markovian, there is no connection $X$--$U$--$Y$ in that path. Similarly, we cannot have $X$--$U$--$U'$, with $U'$ being another exogenous node.  
    Hence, any path must have at least one endogenous node on which we can condition to block it. Now the existence of an active path between $Y$ and $U$ in  $\graph{G}$ would imply an active path between $X$ and $Y$ in $\graph{M}$, hence it cannot exist.
\end{proof}

\begin{proof}[Proof of Theorem~\ref{thm:alg-is-sound}]
We will prove that each case follows from  probability laws and Pearl's do-calculus \citep{causality}.
Thus consider $Y_t$ that satisfies either \ref{cond1a} or \ref{cond1b}.
According to the variable selection rule, either case only occurs if there are no descendants of X in $\mathbf{Y}_t$. 
We thus have  
\begin{align*}
    \label{case1-proof}
    \Pr(\mathbf{y}_t|\Do{x}) &= \Pr(y_t,\mathbf{y}_{t+1}|\Do{x}) \\ 
    &= \Pr(\mathbf{y}_{t+1}|\Do{x})  \Pr(y_t|\Do{x},\mathbf{y}_{t+1}) .
\end{align*}
Rule 3 of do-calculus states that:
\[
\Pr(y_t|\Do{x},\mathbf{y}_{t+1})=\Pr(y_t|\mathbf{y}_{t+1}), 
\]
whenever $Y_t$ and ${X}$ are d-separated by $\mathbf{Y}_{t+1}$ in $\mathcal{G}_{\overline{\mathbf{X}}}$.
Because $\mathbf{Y}_t$ does not contain descendants of $\mathbf{X}$, any path from $Y_t$ to ${X}$ in $\mathcal{G}_{\overline{{X}}}$ goes through some collider which has no descendants in $\mathbf{Y}_{t+1}$.
Now, if $Y_t \in \mathbf{U}$, then $Y_t$ is also d-separated from $\mathbf{Y}_{t+1}$ and $\Pr(y_t|\mathbf{y}_{t+1}) = \Pr(y_t)$. 

Now consider $Y_t$ that satisfies \ref{cond2}. 
Let $\mathbf{Z}'=\pa{Y_t} \setminus \mathbf{Y}_t$; that is, unlike $\mathbf{Z}_t$, $\mathbf{Z}'$ includes $X$ if $X \in \pa{Y_t}$. Similarly, let $\mathbf{y}'$ be $\mathbf{y}_{t+1}$ possibly extended with $x$ if $X \in \pa{Y_t}$.
Denote by $\pi$ an assignment of the parents of $\pa{Y_t}$ that is consistent with $\mathbf{y}'$.
Usual probabilistic manipulation, and d-separation between a node and its nondescendants nonparents given its parents, leads to 
$\Pr(\mathbf{y}_t|\Do{x}) =  \sum_{\mathbf{z}'}  \Pr(y_t|\pi,\Do{x}) \Pr(\mathbf{y}'|\Do{x})$.
Now, because $\Pr(X=x|\mathbf{y}_{t+1},\Do{x})=1$, we obtain 
$\Pr(\mathbf{y}_t|\Do{x}) =   \sum_{\mathbf{z}_{t+1}}  \Pr(y_t|\pi,\Do{x}) \Pr(\mathbf{y}_{t+1}|\Do{x})$. 
Using  the fact that a parent set defines a backdoor set,  the intervened conditional probability of a variable given its parents is identified with its non-intervened probability, which is just the corresponding mechanism. So, 
\begin{align*}
    \Pr(\mathbf{y}_t|\Do{x})  
    &= \sum_{\mathbf{z}_t} \bracket{f_{Y_t}(\mathbf{z}_t,\mathbf{w}_t)=y_t} \Pr(\mathbf{y}_{t+1}|\Do{x}) .
\end{align*}
At last, consider $Y_t$ that satisfies \ref{cond3}. Note that $X$ and $Y_t$ have no common parent (as they are assumed in distinct c-components), and there can be no descendants of $Y_t$ in $\mathbf{Y}_{t+1}$ (because we only add ancestors and we process variables in reverse topological order). Hence, we can apply Lemma~\ref{lem:dsep-for-case4} to show that there is a subset $\mathbf{W}_t$ of the endogenous ancestors of $\mathbf{Y}_{t}$ such that $\mathbf{S}_t$ d-separates $Y_t$ and $X$ in $\graph{G}_{\underline{X}}$, and such that $\mathbf{S}_t \cup \{X\}$ d-separates $Y_t$ and $\mathbf{U}_t$ in $\graph{G}$. 
By probability laws, we have that
$\Pr(\mathbf{y}_t | \Do{x}) =
    \sum_{\mathbf{z}_t} \Pr(y_t|\mathbf{y}_{t+1},\Do{x}) \Pr(\mathbf{y}_{t+1} | \Do{x})$.
It follows from Rule 2 of the do-calculus that 
\[ \Pr(y_t|\mathbf{y}_{t+1},\Do{x})=\Pr(y_t|x,\mathbf{y}_{t+1}) ,
\]
because $Y_t$ and $X$ are d-separated by $\mathbf{Y}_{t+1}$ in the graph obtained by removing edges leaving $X$.
Now since $\mathbf{S}_t \cup \{X\}$ d-separates also $Y_t$ and $U_t$, we can ignore $u_t$ from the right hand side above (if it exists), producing 
\[
    \Pr(\mathbf{y}_t | \Do{x}) = \sum_{\mathbf{z}_t} \widehat{\Pr}(y_t|x,\mathbf{v}_t) \Pr(\mathbf{y}_{t+1} | \Do{x}),
\]
and concluding the proof.
\end{proof}


All the results in this section are still correct when we have multiple interventions in the same c-component, as such a case can essentially be reduced to  a single-intervention case.   

\section{Exploiting  Column Generation}\label{sec:columngen}

We now suppose that a single c-component is intervened upon; that is, $\underline{\Pr}(\mathbf{y}|\Do{x})$ is produced by a linear program as described in the previous section. 
Denote by $\mathbf{C}^*$ the intervened c-component, by $\mathbf{W}^*$ the union of the (endogenous) variables in $\mathbf{C}^*$ and their parents, and by $U^*$ the exogenous variable connected to $\mathbf{C}^*$. For simplicity, we assume all endogenous variables are binary.

There are $2^M$ constraints (\ref{eq:linearconstraint}), where $M := |\mathbf{W}^*|$
(one constraint per configuration of $\mathbf{W}^*$).
Canonicalization may lead to a large cardinality for $U^*$ (one value per possible mechanism), given by 
$\prod_{V \in \mathbf{C}^*} 2^{2^{|\pa{V}|}}$ (Section \ref{section:Background}). 
Depending on the edges among variables in $\mathbf{W}^*$, the cardinality of $U^*$ may be of order $2^{2^M}$.
We assume here that $|\mathbf{W}^*|$ is relatively small, say 4 or 5 variables.
Even then, note that $2^{2^M}$ is already larger than 4 billions for $|\mathbf{W}^*|=5$, hence we should not expect that a direct formulation of the linear program is practical.

We can write down the constraints (\ref{eq:linearconstraint}) in matrix form
as $\mathbf{A} \mathbf{p} = \hat{\mathbf{q}}$, where 
the vector $\mathbf{p}$ contains the optimizing variables $p_u = \Pr(U^*=u)$, indexed by values of $U^*$. And the vector $\hat{\mathbf{q}}$ comes from the input empirical distribution. 
Matrix $\mathbf{A}$ only contains zeros and ones; there are as many rows as constraints, and as many columns as values of $U^*$.

However, only a square matrix is actually needed at any given iteration of the revised simplex algorithm \citep{LP};
that is, we can keep a $2^M\times2^M$ matrix in memory, where each column is defined by a mechanism (hence we only need to find $2^M$ mechanisms). We can thus resort to column generation to sequentially find the relevant columns.

As a digression, note that  \citet{Shridharan2024} have shown that, in some particular cases, the number of columns of $\mathbf{A}$ can be reduced to $2^{|\mathbf{C}^*|2^{|\cup_{V\in\mathbf{C}^*} \pa{V}\backslash\mathbf{C}^*|}}$. 
We assume that those particular cases are treated, whenever they apply and reduce costs, before our techniques are run.

\subsection{Column Generation}

Column generation searches for a column at each iteration of the revised simplex method, by searching $u$ that minimizes the reduced cost \citep{LP}:
\begin{equation}
\label{equation:ReducedCost} 
\gamma_u - \mathbf{d} \cdot \mathbf{a}_u, 
\end{equation}
where $\gamma_u$ is the $u$th coefficient of the objective function,
$\mathbf{d}$ is the current vector of dual costs,
and $\mathbf{a}_u$ is the $u$th column of matrix $\mathbf{A}$. We have used subscripts $u$ because  each coefficient of the objective function, and each column of $\mathbf{A}$, is  fixed when we select a value $u$ of the exogenous variable $U^*$.  
Note that the vector $\mathbf{d}$ is
usually available as a call to the linear solver of choice, so we assume it is available. 
In our context, there are as many dual costs as there are constraints; as constraints in our problem can be indexed by the configuration $\mathbf{w}$ of $\mathbf{W}^*$, we write $\mathbf{d}_\mathbf{w}$ for each entry of $\mathbf{d}$. 
Likewise, the vector $\mathbf{a}_u$ has as many entries as there are constraints; so we write $\mathbf{a}_{u,\mathbf{w}}$ for the entry of $\mathbf{a}_u$ corresponding to the configuration $\mathbf{w}$ of $\mathbf{W}^*$. 

Consider first the term,
\[ 
\mathbf{d} \cdot \mathbf{a}_u = \sum_\mathbf{w} \mathbf{d}_\mathbf{w} \mathbf{a}_{u,\mathbf{w}},
\]
where the summation runs over the values $\mathbf{w}$ of $\mathbf{W}^*$. 
Note that $u$ is not fixed at this point, as we are searching for it;
however, $\mathbf{w}$, for each term in the summation, is fixed.

We start by writing an implicit expression for $\mathbf{a}_u$.
The strategy is to write a generic value $u$ of $U^*$ in binary notation as a sequence of bits;
in fact, as a sequence of blocks of bits, one per mechanism $f_V$.
The procedure is as follows. 
\begin{enumerate}
\item Take each mechanism $f_{V_i}$ in $\mathbf{C}^*$, for $i=1,\dotsc,|\mathbf{C}^*|$. 
\begin{enumerate}
\item If $f_{V_i}$ is a function only of $U$, introduce a single bit $b^i_1 = f_{V_i}(U)$. 
That is, the output of $f_{V_i}(U)$ is simply the bit $b^i_1$ of $U$. 
\item If instead $f_{V_i}$ is a function of $U$ and a set of $n_i$
variables in $\{V_1,\dots,V_{i-1}\}$, then: for {\em each} one of the $2^{n_i}$
configurations of these variables, ordered themselves as binary numbers,
introduce a bit $b^i_j$. Note that this step introduces $2^{n_i}$ bits.
\end{enumerate}
\item Write  
$u=b^1_0\dotsb b^1_{2^{n_1}-1}
\dotsb b^{|\mathbf{C}^*|}_1 \dotsb b^{|\mathbf{C}^*|}_{2^{n_{|\mathbf{C}^*|}}-1}$,
with $n_i=0$ when $f_{V_i}$ does not depend on endogenous variables,
for a generic value of $U^*$.
\end{enumerate}

\begin{example} \label{example:ColumnGenerationBits}
Consider the quasi-Markovian model in Figure  \ref{fig:nonlinear2} (right), 
and focus on the c-component $\mathbf{C}^*$ associated with exogenous variable $U^*=U_1$.
Order the variables in $\mathbf{W}^*$, $W$, $X$ and $Z$, 
using lexicographic order ($V_1$ is $W$, $V_2$ is $X$, $V_3$ is $Z$).
Now examine mechanisms associated with variables in $\mathbf{C}^*$.
Mechanism $f_X$ depends only on $U_1$, hence we write:
$f_X(u)=b^2_0$.
Mechanism $f_W$ depends on $X$ and $Z$ (in this order) besides $U_1$, so we must code 
the four configurations of $X,Z$, ordered using binary notation: 
\begin{equation}
\label{equation:CodeFW}
\begin{array}{ll}
f_W(0,0,u)=b^1_0, & f_W(0,1,u)=b^1_1, \\ 
f_W(1,0,u)=b^1_2, & f_W(1,1,u)=b^1_3. 
\end{array}
\end{equation}
Thus a generic value $u$ of $U_1$ is written
as $b^1_0 b^1_1 b^1_2 b^1_3 b^2_0$. 
Consequently, there are $2^5 = 32$ values of $U$, 
agreeing with  $\prod_{V \in \mathbf{C}^*} 2^{2^{|\pa{V}|}} = 2^{2^0} \times 2^{2^2}$. 
$\Box$
\end{example}
  
We   use these bits to build a ``symbolic'' version of each $\mathbf{a}_{u,\mathbf{w}}$, so
as to leave $\mathbf{a}_u$ as a function of the sought for value $u$:
\begin{enumerate}
\item For each variable $V_i$ in $\mathbf{C}^*$, take   mechanism $f_{V_i}$ and:
\begin{enumerate}
\item get the bit $b^i_j$ that corresponds to the value of $f_{V_i}$ evaluated at $(u,\mathbf{w})$;
\item if $v_i=1$, then insert $b^i_j$ into a list $L^+$;
\item if $v_i=0$, then insert $(1-b^i_j)$ into a list $L^-$.
\end{enumerate}
\item Return  
$\mathbf{a}_{u,\mathbf{w}}=\prod_{b^+ \in L^+} b^+ \prod_{b^- \in L^-} (1-b^-)$.
\end{enumerate}
Note that each $b^+$ and $b^-$ is restricted to be integer in $\{0,1\}$.

We must now eliminate the products of bits; we do so resorting to linear   constraints. 
More precisely, we replace the product in the last step of the previous algorithm by the following set of $|\mathbf{C}^*| +1$ linear constraints:
\[
0 \leq \mathbf{a}_{u,\mathbf{w}} \leq b^+, \quad 
\mbox{ for each } b^+ \in L^+,
\]
\[
0 \leq \mathbf{a}_{u,\mathbf{w}} \leq (1-b^-), \quad 
\mbox{ for each } b^- \in L^-,
\]
\[
1 - |\mathbf{C}^*| + \sum_{b^+ \in L^+} b^+ + \sum_{b^- \in L^-} (1-b^-) \leq \mathbf{a}_{u,\mathbf{w}} \leq 1.
\] 

\begin{example}\label{ColumnGenerationColumn}
Consider again Figure \ref{fig:nonlinear2} (right),
where each $u$ is written as bits
$b^1_0 b^1_1 b^1_2 b^1_3 b^2_0$.
A generic column $\mathbf{a}_u$ has an entry
per configuration $\mathbf{w} = (wxz)$ of $\mathbf{W}^*$,
as follows:  
{\em if} $w = b^1_{2x+z}$, {\em and} 
$x = b^2_0$, {\em then} 
$\mathbf{a}_{u,\mathbf{w}} = 1$; 
{\em otherwise},
$\mathbf{a}_{u,\mathbf{w}} = 0$.
Abusing notation (equating ``true'' and $1$, ``false'' and $0$), we  write:
\[
\mathbf{a}_{u,\mathbf{w}} = 
  (b^1_{2x+z} \leftrightarrow w) 
   \wedge (b^2_0 \leftrightarrow x).
\]
For binary variable $V$ and any bit $b$, $(v \leftrightarrow b)$ is just $b$ when $v=1$, and is just $(1-b)$ when $v=0$. Moreover, with binary variables we can reproduce   conjunction using product.
For instance if $wxz=000$, we have $\mathbf{a}_{u,000}=(1-b^1_0) (1-b^2_0)$.
That is, each element of the column $\mathbf{a}_{u,\mathbf{w}}$ is a product of bits (or negated bits) of $u$. 
Additional details can be found in the Supplementary Material (Section \ref{section:SupplementaryExample}).
$\Box$
\end{example}

To actually process the reduced cost (Expression (\ref{equation:ReducedCost})) within column generation, we still need to write $\gamma_u$, the coefficient of $\Pr(U^*=u)$ in the objective function, as a function of (the bits of) $u$. 
Note that $\gamma_u = \Pr(\mathbf{y}|\Do{X=x},U^*=u)$.
As described in Section \ref{section:Algorithm}, our proposed algorithm generates $\gamma_u$ sequentially;
in the end, $\gamma_u$ is a summation with as many terms as there are configurations of $\mathbf{W}^*\backslash\{X\}$. Moreover, each term is a product of $|\mathbf{C}^*|-1$ bits (or negated bits) in the binary notation for $U^*$ (the same binary encoding discussed previously), as all mechanisms in $\mathbf{C}^*$ contribute, except the mechanism of the intervened variable.  
%

We can thus use the same techniques discussed previously to code $\mathbf{a}_u$ to build $\gamma_u$ as a function of the bits of $U^*$.  
An example should clarify the idea.

\begin{example}\label{example:ColumnGenerationObjective}
Consider again the quasi-Markovian model in  Figure \ref{fig:nonlinear2} (right). 
By running the algorithm in the previous section, we obtain:
$\Pr(y|\Do{x}) = \sum_{u} \gamma_{u} \Pr(u)$, where 
\[
\gamma_{u} =
\textstyle \sum_{w,z} \widehat{\Pr}(y|w,x,z)
\bracket{f_W(x,z,u)=w}
\widehat{\Pr}(z|x).
\]
Recall that 
$\bracket{f_W(x,z,u)=z}$ is given by Expression (\ref{equation:CodeFW}).
If we want $\Pr(Y=y|\Do{X=1})$; then $\gamma_u$ is equal to:
\[
\begin{array}{l}
\widehat{\Pr}(y|W\!=\!0,X\!=\!1,Z\!=\!0)\widehat{\Pr}(Z\!=\!0|X\!=\!1) (1-b^1_2) + \\
\widehat{\Pr}(y|W\!=\!0,X\!=\!1,Z\!=\!1)\widehat{\Pr}(Z\!=\!1|X\!=\!1) (1-b^1_3) + \\
\widehat{\Pr}(y|W\!=\!1,X\!=\!1,Z\!=\!0)\widehat{\Pr}(Z\!=\!0|X\!=\!1)  b^1_2 + \\
\widehat{\Pr}(y|W\!=\!1,X\!=\!1,Z\!=\!1)\widehat{\Pr}(Z\!=\!1|X\!=\!1)  b^1_3,  
\end{array}
\]
a function of the bits of $U_1$. In a more complex example we might have to handle products of bits in the objective function, by adding linear integer constraints as before.
$\Box$
\end{example}

By writing $\gamma_u$ and $\mathbf{a}_u$ as depending on bits of $U^*$, we can find a value $u$ that minimizes Expression (\ref{equation:ReducedCost}). By iterating this process as usual in any implementation of column generation \citep{LP}, we can reach $\underline{\Pr}(\mathbf{y}|\Do{x})$ as desired.

\subsection{Experiments}

We have implemented the algorithm in the previous section,\footnote{It will be made publicly available if the paper is accepted.} relying on the Gurobi solver for master and auxiliary programs.\footnote{Gurobi version 12.0.2 (Linux): www.gurobi.com.}
We  report representative experiments here.  

We start with the quasi-Markovian SCM depicted in Figure~\ref{fig:nonlinear2} (right). This SCM is based on a practical problem faced by one of the authors; namely, the evaluation of causes in a low-latency service pipeline. Binary treatment $X$ signals activation of a newly deployed AI system. High processing requests are indicated by $Z$ (whether or not average load exceeds, say, 80\%); $Z$ mediates the effect of $X$ on tail latency $W$, and the latter may in turn trigger an incident indicated by $Y$. We also allow for a direct path from $X$ to $W$, caused by database calls issued by the AI model. There are latent pressures on the pipeline: $U_1$ indicates heavy traffic induced by marketing campaigns, and $U_2$ refers to the degradation of an external API that affects both $Z$ and $Y$. 

\begin{figure}[t]
    \centering
   \begin{tikzpicture}[>=latex]
      \tikzstyle{endo} = [draw,circle,fill=black,minimum size=4pt,inner sep=0.0em]
      \tikzstyle{exo} = [draw,circle,fill=white,minimum size=4pt,inner sep=0.0em]
      \node[exo,label=left:$U_1$] (U1) at (0,1) {};
      \node[exo,label=right:$U_2$] (U2) at (4,-1) {}; 
      \node[endo,label=below:$Z_1$] (B1) at (0,-1) {};
      \node[endo,label=right:$Y$] (Y) at (5,0) {};
      \node[endo,label=left:$X$] (X) at (-1,0) {};
      \node[endo,label=above:$W_1$] (A1) at (1,0) {};
      \node[font=\large] (anchorL) at (2,0) {$\dots$};
      \node[] (anchorR) at (2.15,0) {};
      \node[endo,label=above:$W_N$] (An) at (3,0) {};
      \node[endo,label=below:$Z_M$] (Bm) at (2,-1) {};
      \node[font=\large] (dots) at (1,-1) {$\dots$};
      \foreach \x/\y in {X/B1,B1/A1,U1/X,U1/A1,U2/Y,X/A1,A1/anchorL,anchorR/An,An/Y,U1/An,Bm/A1,Bm/An,U2/Bm,X/Bm/,B1/An} {
        \draw[->] (\x) edge (\y);
      };/
      \draw[->] (U2) to [out=165,in=15] (B1);      
    \end{tikzpicture}
    \caption{A parameterized expansion of the graph in Figure \ref{fig:nonlinear2} (right). We have $|\mathbf{C}^*|=N+1$ and $|\mathbf{W}^*|=N+M+1$.}
    \label{figure:SeriesIntervention}
    \end{figure}
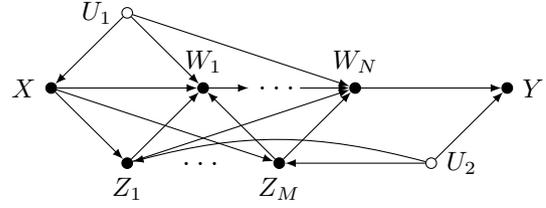

We also consider a parameterized version of this SCM, where there are $N$ variables $W_i$ in the direct path between $X$ and $Y$, all of them in the c-component connected with $U_1$, and $M$ observed confounders $Z_j$, children of $U_2$ and  parents of all $W_i$. The template is depicted in Figure \ref{figure:SeriesIntervention}.
For $N=M=1$ we obtain the graph in Figure \ref{fig:nonlinear2} (right). No pair $(M,N)$ can be handled by techniques by \cite{sachs}.
Additional details can be found in the Supplementary Material (Section \ref{section:Series}).

We wish to compute bounds on $\Pr(y|\Do{X=x})$. There are on the order of $2^{N2^M}$ optimization variables in the corresponding linear program. To run column generation, we build auxiliary linear integer programs that produce values of $U_1$ coded as $b^1_0...b_{2^{M+1}-1}^1...b_{2^{M+1}-1}^Nb^{N+1}_0$,
where each block of bits $b^i_0...b_{2^{M+1}-1}^i$ corresponds to a variable $W_i$ and bit
$b^{N+1}_0$ corresponds to $X$.

Table \ref{table:Results} shows the comparison between the execution times with our column generation scheme (CG) and with just the direct linear program (LP) conveyed by Expression (\ref{equation:MixedFactorization}).
The latter approach cannot handle several cases; as can be seen in the table, the running time of LP grows dramatically and is, except for small problems, much larger than the running time of CG (all tests run in an AMD 32-cores machine). A particularly striking case is $M=3$, $N=1$, where CG is 20.000 times faster than LP.

\subsection{Bonus: Solution by single program}

We have explored, in the previous section, an encoding for a {\em single} coefficient $\gamma_u$ and a {\em single} column $\mathbf{a}_u$. However, we know that to optimize $\Pr(\mathbf{y}|\Do{x})$ we must only find $M$ such pairs coefficient/column --- if we simultaneously find the $M$ right columns (instead of one), we can obtain the desired probability bounds. 

This leads to the following strategy: \\
{\bf 1)} Build $M$ copies of the bits described in the previous section: for each bit $b^i_j$, we have $M$ bits $b^{i,k}_{j}$. \\
{\bf 2)} Then use, for each $k\in\{1,\dots,M\}$, the bits $b^{i,k}_{j}$ to build constraints for a column $\mathbf{a}_{u_k}$. So, we have $M$ columns that depend on the bits. \\
{\bf 3)} Similarly, use, for each $k\in\{1,\dots,M\}$, the bits $b^{i,k}_j$ to build a coefficient $\gamma_{u_k}$. So, we have $M$ coefficients that depend on the bits. \\
{\bf 4)} Write down a single objective function as a sum $\sum_k \gamma_{u_k} \Pr(u_k)$ subject to $\mathbf{A} \mathbf{p} = \hat{\mathbf{q}}$ where $\mathbf{A}$ is now the square matrix with the $M$ columns just built, $\mathbf{p}$ is a vector with the optimizing values $\Pr(u_{k})$, and $\hat{\mathbf{q}}$ is defined as before. 
These expressions contains products of the form $m p$, where $m\in\{0,1\}$ and $p$ is a real (both optimizing variables). Replace each such product by a fresh variable $\alpha_{m,p}$ subject to $0\leq\alpha_{m,p}\leq m$ and $p+m-1 \leq \alpha_{m,p}\leq p$. \\
The minimum/maximum for the latter linear integer program is exactly the lower/upper bound we desire. 

This strategy has been employed in probabilistic logic \citep{Ianni2015}. The advantage of a non-iterative strategy is simplicity of implementation. Past experience suggests that column generation tends to be faster, an issue to be verified empirically in our present context.

\begin{table}[t]
\centering
\begin{tabular}{|c|c|c|c|} \hline
$M$ & $N$ & CG(s) & LP(s) \\ \hline  
1 & 1 &  0.749& 0.661\\   
1 & 2 & 0.271 & 1.55 \\  
1 & 3 &  1.81 & 106 \\  
1 & 4 &  16.0&  411\\  
1 & 5 &  1050& 7174 \\  
2 & 1 & 0.216 & 4.65 \\  \hline
\end{tabular}
\begin{tabular}{|c|c|c|c|} \hline
$M$ & $N$ & CG(s) & LP(s) \\ \hline  
2 & 2 & 0.867 & 2494 \\ 
2 & 3 &94.5 & - \\  
2 & 4 & 4360 & - \\  
3 & 1 & 0.207 & 4207 \\  
3 & 2 & 6.64 & - \\ 
3 & 3 &  3138 & - \\ \hline 
\end{tabular}
\caption{Runs of Column Generation (CG) and direct Linear Programming (LP),
in seconds, for several pairs $M,N$. Entries marked ``-'' mean that the LP
solver did not finish within 3 hours of execution.
}
\label{table:Results}
\end{table}

\section{Conclusion}
\label{section:Conclusion}

In this paper we have investigated the computation of probability bounds for quasi-Markovian SCMs subject to interventions. We described a shorter proof for \citet{shridharan23icml}'s key reduction to multilinear programming.
We then proposed a new algorithm that exploits the presence of input probabilities when building linear programs   in the presence of a single intervention. 
We presented a column generation scheme to solve such linear programs, and described an approach that computes lower or upper bounds using a single linear integer program. Our experiments showed that column generation offers significant improvements over direct linear programming. 

Future work should extend our results to multiple interventions.
It is also important to characterize the complexity of our algorithm, perhaps by connecting it with graph-theoretical quantities, and to explore tree decompositions or similar data structures to accelerate the construction of linear programs.
Another promising avenue is to combine column generation with special cases that already reduce linear programs; for example, the cases discussed by \cite{Shridharan2024}. 

Looking forward, it will be valuable to combine binary (and categorical) variables with continuous ones, so as to expand practical application. That extension will likely require different techniques to handle continuous variables, even if continuous variables are restricted to say Gaussian distributions. 



\section*{Acknowledgements}

We thank the Center for Artificial Intelligence at Universidade de São Paulo (C4AI-USP), with support by the S\~ao Paulo Research Foundation (FAPESP grant 2019/07665-4) 
and by the IBM Corporation.
We thank the Instituto de Ciência e Tecnologia Itaú (ICTi), for providing key funding
for this work through the Centro de Ciência de Dados (C2D) at Universidade de São Paulo.
D.D.M.\ was partially supported by CNPq grant 305136/2022-4 and FAPESP grant 2022/02937-9.
F.G.C.\ was partially supported by CNPq grants 312180/2018-7 and 305753/2022-3. 
The authors also thank support by CAPES - Finance Code 001.

\bibliography{refs}

\newpage

\onecolumn

\title{Multilinear and Linear Programs for Partially Identifiable 
Queries in quasi-Markovian Structural Causal Models \\(Supplementary Material)}
\maketitle

\vspace*{2cm}

\section{Additional Details about Example \ref{ColumnGenerationColumn}}\label{section:SupplementaryExample}

Here we offer details about Example \ref{ColumnGenerationColumn}. 
We start by repeating the relevant graph (left) and presenting the corresponding 
intervened semi-marginal graph for $\Do{X=x}$ (right):

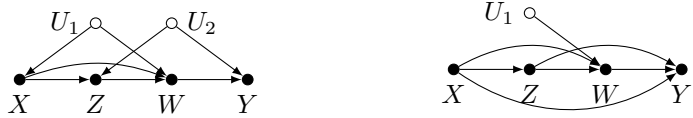
\begin{center}
    \begin{tikzpicture}[>=latex]
      \tikzstyle{endo} = [draw,circle,fill=black,minimum size=4pt,inner sep=0.0em]
      \tikzstyle{exo} = [draw,circle,fill=white,minimum size=4pt,inner sep=0.0em]
      \node[exo,label=left:$U_1$] (U1) at (0,.75) {};
      \node[exo,label=right:$U_2$] (U2) at (1,.75) {}; 
      \node[endo,label=below:$Z$] (Z) at (0,0) {};
      \node[endo,label=below:$Y$] (Y) at (2,0) {};
      \node[endo,label=below:$X$] (X) at (-1,0) {};
      \node[endo,label=below:$W$] (W) at (1,0) {};
      \foreach \x/\y in {X/Z,Z/W,W/Y,U1/X,U1/W,U2/Z,U2/Y} {
        \draw[->] (\x) edge (\y);
      };
      \draw[->] (X) to [out=20,in=160] (W);
    \end{tikzpicture}
    \hspace*{2cm}
    \begin{tikzpicture}[>=latex]
      \tikzstyle{endo} = [draw,circle,fill=black,minimum size=4pt,inner sep=0.0em]
      \tikzstyle{exo} = [draw,circle,fill=white,minimum size=4pt,inner sep=0.0em]
      \node[exo,label=left:$U_1$] (U1) at (1,.75) {};
      \node[endo,label=below:$X$] (A) at (0,0) {};
      \node[endo,label=below:$Z$] (B) at (1,0) {};
      \node[endo,label=below:$W$] (C) at (2,0) {};
      \node[endo,label=below:$Y$] (D) at (3,0) {};
      \foreach \x/\y in {U1/C,A/B,B/C,C/D} {
        \draw[->] (\x) edge (\y);
      }; 
      \draw[->] (B) to [bend left] (D);
      \draw[->] (A) to [bend left] (C);      
      \draw[->] (A) to [out=-35,in=-145] (D);      
    \end{tikzpicture} 
\end{center}

We take all conventions described in Example \ref{example:ColumnGenerationBits}.
That is, variables in $\mathbf{W}^*$ are ordered so that
$V_1$ is $W$, $V_2$ is $X$, $V_3$ is $Z$; 
values of $U_1$ are written in binary notation as $b^1_0 b^1_1 b^1_2 b^1_3 b^2_0$;
and we code mechanisms $f_W(X,Z,U_1)$ and $f_X(U_1)$ as follows:
\[
f_W(0,0,u)=b^1_0, \quad f_W(0,1,u)=b^1_1, \quad 
f_W(1,0,u)=b^1_2, \quad f_W(1,1,u)=b^1_3, \quad 
f_X(u)=b^2_0.
\]
As noted in  Example \ref{ColumnGenerationColumn}, 
a generic column $\mathbf{a}_u$ has a entry
per configuration $\mathbf{w} = (wxz)$ of $\mathbf{W}^*$,
\[
\mathbf{a}_{u,\mathbf{w}} = 
  (b^1_{2x+z} \leftrightarrow w) 
   \wedge (b^2_0 \leftrightarrow x).
\]
The following table explicitly shows the entries of $\mathbf{a}_u$:
\begin{center}
\begin{tabular}{|c|c|}   \hline
$w=(wxz)$ & $\mathbf{a}_u$, 
 for $u= b^1_0 b^1_1 b^1_2 b^1_3 b^2_0$ \\ \hline \hline
000 & $  (b^1_0 \leftrightarrow 0) \wedge  (b^2_0 \leftrightarrow 0)  $ \\ \hline
001 & $  (b^1_1 \leftrightarrow 0) \wedge  (b^2_0 \leftrightarrow 0)  $ \\ \hline
010 & $  (b^1_2 \leftrightarrow 0) \wedge  (b^2_0 \leftrightarrow 1)  $ \\ \hline
011 & $  (b^1_3 \leftrightarrow 0) \wedge  (b^2_0 \leftrightarrow 1)  $ \\ \hline
100 & $  (b^1_0 \leftrightarrow 1) \wedge  (b^2_0 \leftrightarrow 0)  $ \\ \hline
101 & $  (b^1_1 \leftrightarrow 1) \wedge  (b^2_0 \leftrightarrow 0)  $ \\ \hline
110 & $  (b^1_2 \leftrightarrow 1) \wedge  (b^2_0 \leftrightarrow 1)  $ \\ \hline
111 & $  (b^1_3 \leftrightarrow 1) \wedge  (b^2_0 \leftrightarrow 1)  $ \\ \hline
\end{tabular}
\end{center}

For binary variable $V$ and any bit $b$, $(v \leftrightarrow b)$ is just $b$ when $v=1$, and is just $(1-b)$ when $v=0$. Moreover, with binary variables we can reproduce   conjunction using product.
For instance if $wxz=000$, we have $\mathbf{a}_{u,000}=(1-b^1_0) (1-b^2_0)$.
That is, each element of the column $\mathbf{a}_{u,\mathbf{w}}$ is a product of bits (or negated bits) of $u$. 
Hence we have:

\begin{center}
\begin{tabular}{|c|c|}   \hline
$w=(wxz)$ & $\mathbf{a}_u$, 
 for $u= b^1_0 b^1_1 b^1_2 b^1_3 b^2_0$ \\ \hline \hline
000 & $  (1-b^1_0)(1-b^2_0)$ \\ \hline
001 & $  (1-b^1_1) (1-b^2_0)$ \\ \hline
010 & $  (1-b^1_2)b^2_0$ \\ \hline
011 & $  (1-b^1_3)b^2_0$ \\ \hline
100 & $  b^1_0(1-b^2_0)$ \\ \hline
101 & $  b^1_1(1-b^2_0)$ \\ \hline
110 & $  b^1_2b^2_0$ \\ \hline
111 & $  b^1_3b^2_0$ \\ \hline
\end{tabular}
\end{center}

Note that all bits $b^i_j$ are integer variables in $\{0,1\}$.
Then each product of bits can be turned into linear integer constraints;
here are a few instances: 
\begin{itemize}
\item $\mathbf{a}_{u,000}$ (that is, $wxz=000$):
\[
0 \leq \mathbf{a}_{u,000} \leq (1-b^1_0), \quad
0 \leq \mathbf{a}_{u,000} \leq (1-b^2_0), \quad
1-b^1_0-b^2_0 \leq \mathbf{a}_{u,000} \leq 1.
\]
\item $\mathbf{a}_{u,011}$ (that is, $wxz=011$):
\[
0 \leq \mathbf{a}_{u,011} \leq (1-b^1_3), \quad
0 \leq \mathbf{a}_{u,011} \leq b^2_0, \quad
 b^2_0 -b^1_3 \leq \mathbf{a}_{u,011} \leq 1.
\]
\item $\mathbf{a}_{u,111}$ (that is, $wxz=111$):
\[
0 \leq \mathbf{a}_{u,111} \leq b^1_3, \quad
0 \leq \mathbf{a}_{u,111} \leq b^2_0, \quad
b^1_3 + b^2_0 -1 \leq \mathbf{a}_{u,111} \leq 1.
\]
\end{itemize}
Hence the term $\mathbf{d} \cdot \mathbf{a}_u$ in the reduced cost (Expression (\ref{equation:ReducedCost})) is  a summation with 8 terms, one per configuration of $(wxz)$, subject to 24 linear constraints and the fact that all $b^i_j \in \{0,1\}$.  

As described in Example \ref{example:ColumnGenerationObjective}, we can use the same sort of encoding via bits for the objective function; in this particular example the objective function is itself a linear function of bits. A more complex example may exhibit products of bits (or negated) bits in the objective function. 

\begin{figure}
\begin{center}
    \begin{tikzpicture}[>=latex]
      \tikzstyle{endo} = [draw,circle,fill=black,minimum size=4pt,inner sep=0.0em]
      \tikzstyle{exo} = [draw,circle,fill=white,minimum size=4pt,inner sep=0.0em]
      \node[exo,label=above:$U_3$] (U3) at (0.5,.75) {};
      \node[exo,label=above:$U_2$] (U2) at (4,.75) {};
      \node[exo,label=above:$U_1$] (U1) at (1.5,.75) {};
      \node[endo,label=below:$S$] (S) at (0,0) {};
      \node[endo,label=below:$Z$] (Z) at (1,0) {};
      \node[endo,label=below:$X$] (X) at (2,0) {};
      \node[endo,label=below:$W$] (R) at (3,0) {};
      \node[endo,label=below:$Y$] (Y) at (4,0) {};
      \node[endo,label=above:$T$] (T) at (2.5,0.75) {};
      \foreach \x/\y in {S/Z,X/Z,X/R,R/Y,X/T,T/Y} {
        \draw[->] (\x) edge (\y);
      };
      \foreach \x/\y in {U1/X,U1/R,U1/T,U2/Y,U3/S,U3/Z} {
        \draw[->] (\x) edge (\y);
      };
      \draw[->] (Z) to [out=-50,in=-130] (R);
    \end{tikzpicture} 
    \hspace*{1cm}
        \begin{tikzpicture}[>=latex]
      \tikzstyle{endo} = [draw,circle,fill=black,minimum size=4pt,inner sep=0.0em]
      \tikzstyle{exo} = [draw,circle,fill=white,minimum size=4pt,inner sep=0.0em]
      \node[exo,label=above:$U_1$] (U1) at (1.5,.75) {};
      \node[endo,label=below:$S$] (S) at (0,0) {};
      \node[endo,label=below:$Z$] (Z) at (1,0) {};
      \node[endo,label=below:$X$] (X) at (2,0) {};
      \node[endo,label=below:$W$] (R) at (3,0) {};
      \node[endo,label=below:$Y$] (Y) at (4,0) {};
      \node[endo,label=above:$T$] (T) at (2.5,0.75) {};
      \foreach \x/\y in {S/Z,X/Z,X/R,R/Y,X/T,T/Y} {
        \draw[->] (\x) edge (\y);
      };
      \foreach \x/\y in {U1/R,U1/T} {
        \draw[->] (\x) edge (\y);
      };
      \draw[->] (Z) to [out=-50,in=-130] (R);
    \end{tikzpicture}
\end{center}
\caption{Left: a quasi-Markovian model. Right: the intervened semi-marginal graph for $\Do{X=x}$.}
\label{figure:ExtendedRunningExample}
\end{figure}
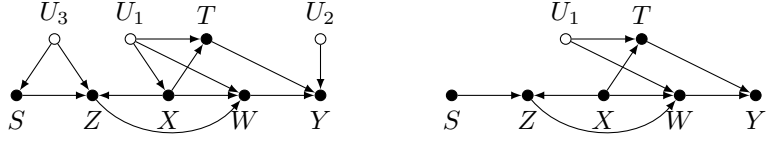

To illustrate the latter possibility, consider the more complex graph 
in Figure \ref{figure:ExtendedRunningExample} (left).
We focus on the expression for the objective function, where the goal 
is to bound $\Pr(y|\Do{X=x})$. Hence we have $\mathbf{C}^* = \{T,W,X\}$ 
and $\mathbf{W}^*=\{T, W, X, Z\}$. We adopt this lexicographic order 
of variables. We now write a value $u$ of $U_1$ as a sequence of bits
$b^1_0 b^1_1 b^2_0 b^2_1 b^2_2 b^2_3 b^3_0$, where the bits are associated 
with mechanisms $f_T(X,U_1)$, $f_W(X,Z,U_1)$ and $f_X(U_1)$ are as follows:
\[
f_T(0,u) = b^1_0, \; f_T(1,u) = b^1_1, \;
f_W(0,0,u)=b^2_0, \; f_W(0,1,u)=b^2_1, \; 
f_W(1,0,u)=b^2_2, \; f_W(1,1,u)=b^2_3, \; 
f_X(u)=b^3_0.
\]
The algorithm in Section \ref{section:Algorithm}  produces 
\[
\Pr(Y=y|\mathrm{do}(X=1))=\sum_u\gamma_u\Pr(U_1=u),
\]
with (introducing a few auxiliary variables $\gamma_{u,\cdot}$):
\begin{eqnarray*}
\gamma_u & = &  \widehat{\Pr}(y|W=0,T=0) \gamma_{u,1} +
           \widehat{\Pr}(y|W=0,T=1) \gamma_{u,2} + \\
& &            \widehat{\Pr}(y|W=1,T=0) \gamma_{u,3} +
           \widehat{\Pr}(y|W=1,T=1) \gamma_{u,4},           
\end{eqnarray*}
\[
\gamma_{u,1} = \widehat{\Pr}(Z=0|X=1) (1-b^1_1)(1-b^2_2) + 
              \widehat{\Pr}(Z=1|X=1) (1-b^1_1)(1-b^2_3),
\]
\[
\gamma_{u,2} = \widehat{\Pr}(Z=0|X=1) b^1_1(1-b^2_2) + 
              \widehat{\Pr}(Z=1|X=1) b^1_1(1-b^2_3),
\]
\[
\gamma_{u,3} = \widehat{\Pr}(Z=0|X=1) (1-b^1_1)b^2_2 + 
              \widehat{\Pr}(Z=1|X=1) (1-b^1_1)b^2_3,
\]
\[
\gamma_{u,4} = \widehat{\Pr}(Z=0|X=1) b^1_1 b^2_2 + 
              \widehat{\Pr}(Z=1|X=1) b^1_1 b^2_3.
\]
We can then collect all these expressions into a single one (as done in previous examples), or feed them separately to the appropriate linear solver. In any case, the key point here is that we have to deal with terms such as $b'b''$ or $b'(1-b'')$ or $(1-b')(1-b'')$, where $b'$ and $b''$ are bits that appear in the program as integer variables in $[0,1]$. 
There are three cases:
\begin{itemize}
\item To handle $b'b''$, introduce a fresh optimization variable $\beta$ and constraints
\[
0 \leq \beta \leq b', \quad 0 \leq \beta \leq b'', \quad b'+b''-1\leq\beta\leq 1.
\]
\item To handle $b'(1-b'')$, introduce a fresh optimization variable $\beta$ and constraints
\[
0 \leq \beta \leq b', \quad 0 \leq \beta \leq 1-b'', \quad b'-b''\leq\beta\leq 1.
\]
\item To handle $(1-b')(1-b'')$, introduce a fresh optimization variable $\beta$ and constraints
\[
0 \leq \beta \leq 1-b', \quad 0 \leq \beta \leq 1-b'', \quad 1-b'-b''\leq\beta\leq 1.
\]
\end{itemize}

\section{Additional Details about Experiments}\label{section:Series}

We start by providing more details on the characteristics of the SCMs depicted in Figure \ref{figure:SeriesIntervention}.
We wish to compute bounds on $\Pr(y|\Do{X=x})$. Note that there are on the order of $2^{N2^M}$ optimization variables in the corresponding linear program containing values of $\Pr(U_1)$. 

Write $U_1$ as $b^1_0...b_{2^{M+1}-1}^1...b_{2^{M+1}-1}^Nb^{N+1}_0$,
where each block $b^i_0...b_{2^{M+1}-1}^i$ corresponds to a variable $W_i$ and 
$b^{N+1}_0$ corresponds to $X$.
The subindex corresponds to the binary number associated with the vector $(W_{i-1},Z_1,...,Z_M)$,
where $Z_M$ is the least significant bit. 

\subsection{Master Linear Program}

\subsubsection{Objective function}

Using the algorithm in Section \ref{section:Algorithm}, we get the objective function:
\begin{equation}\label{eq:scalableObjMaster}
   P(y|\Do{X=x}) =  
            \sum_{
           W_i,Z_j,U} P(y|W_N)P(Z_M,...,Z_1|x)
           \prod_{i=1}^NP(W_i|W_{i-1},Z_M,...,Z_1,U)P(U).
\end{equation}

\subsubsection{Constraints} 
    
We have:
\begin{equation}\label{eq:scalableConstrMaster}
\sum_{u \mapsto (W_1,W_{2},...,W_N,X,Z_1,...,Z_M)} P(u) = P(W_N,W_{N-1},...,W_1|X,Z_1,...,Z_M)P(x).
\end{equation}

\subsection{Auxiliary Linear Integer Program}

\subsubsection{The Costs}
    
Note that in Equation \ref{eq:scalableObjMaster}, each term of the form $P(W_{i}|W_{i-1},Z_1,...,Z_M,U)$ can be expressed as a function of bits from $U_1$, more specifically: 
    \begin{equation}
 P(W_{i}|W_{i-1},Z_M,...,Z_1,U) =  b^{i*}_{Z_M+2Z_{M-1}+...2^{M-1}Z_1+2^MW_{i-1}}.
    \end{equation}
    Let us call $Z_M+2Z_{M-1}+...2^{M-1}Z_1+2^MW_{i-1} = w_i$, in which $w_i$ is the binary number associated with the vector $(W_{i-1},Z_1,...,Z_M)$, i.e. the realization for parents of $W_i, i \in \{1,2,...N \}$ (adopt $W_{0} =  X$). Furthermore,   define:
    \begin{equation}\label{eq:starDefintiom}
        b^{i*}_{w_i} = \begin{cases}
            b^i_{w_i}, W_i = 1, \\
            1-b^i_{w_i}, W_i = 0.
        \end{cases}
    \end{equation}
    Therefore the cost can be written as:
    \begin{equation}\label{eq:cost}
       \begin{aligned}
           \gamma_u = \sum_{Z_j,W_i}P(y|Z_N)P(Z_M,...,Z_1|x)\prod_i b^{i*}_{w_i}.
       \end{aligned} 
    \end{equation}
Now we define $\prod_i b^{i*}_{w_i} = \beta_{x,w}$, where $w$ is the binary number associated with $(W_1,...,W_{N},X,Z_1,...,Z_M)$, i. e. the c-component and it's tail realization (note that the subindex $x$ is redundant, however we leave it for emphasis). The reason for this will be evident in the next section.

\subsubsection{The Columns}

Each column $a_u$ can be indexed by $w$ for a specific realization and it's value can be expressed as: $a_{u,w} = b_0^{N+1*}\beta_{x,w}$, which gives:
\begin{equation}
    a_{u,w} = \begin{cases}
        b_{0}^{N+1}\beta_{1,w}, x =1, \\
        (1-b_{0}^{N+1})\beta_{0,w}, x = 0.
    \end{cases}
\end{equation}

\subsubsection{The Linear Program}

We write a linear program in optimizing variables $\beta_{0,w},\beta_{1,w},a_{u,w}\ , w \in \{0,1,..., 2^{M+N+1}-1\}$, a set with $\mathcal{O}(2^{M+N})$ optimizing variables. Restrictions:
\begin{itemize}
    \item $\beta_{x,w}$:
    \begin{equation}\label{eq:integerProgramingBeta}
        \begin{aligned}
        &0 \leq \beta_{x,w} \leq b^{i*}_{w_i}, \forall i \in \{1,2...,N\}, \\
        &1 - N + \sum_i b^{i*}_{w_i}\leq \beta_{x,w} \leq 1.
        \end{aligned}
    \end{equation}
\item $a_{u,w}$:
\begin{equation}\label{eq:integerProgramingColumn}
        \begin{aligned}
        &0 \leq a_{u,w} \leq b^{N+1*}_{0}, \\
        & 0 \leq a_{u,w} \leq \beta_{x,w},\\
        &- 1 +b^{N+1*}_{0} + \beta_{x,w}+\leq a_{u,w} \leq 1.
        \end{aligned}
    \end{equation}
\end{itemize}
Note that there  are $\mathcal{O}(2^{M+N})$ constraints. 

The objective function is:
\begin{equation}\label{eq:subproblemObj}
    \begin{aligned}
    &\sum_{w,X = 1} P(y|W_N)P(Z_M,...,Z_1|x)x_0\beta_{1,w} + \\
    &\sum_{w,X= 0} P(y|W_N)P(Z_M,...,Z_1|x)(1-x_0)\beta_{0,w} - \\
    &\sum_{w} d_wa_{u,w}.
\end{aligned}
\end{equation}
 This ensures that, for the $\beta_{x,w}$ variables, only $\beta_{x_0,w}$ has non-zero coefficient in the objective function.

\subsubsection{Initialization}

To obtain the first dual cost vector, we need an initial value base so we can start the column generation procedure. In our implementation we used the bigM method \cite{LP}, with $M = 10^4$, with an initial basis $I_{2^{M+N+1}+1}$, i. e., the identity matrix in the size of the restrictions.

\end{document}